%% file: paper.tex
\documentclass[11pt,letterpaper]{article}
\usepackage{emnlp2017}
\usepackage{times}
\usepackage{latexsym}

\emnlpfinalcopy

\def\emnlppaperid{392}

\usepackage{url}

\usepackage{booktabs}       
\usepackage{amsfonts}       
\usepackage{nicefrac}       
\usepackage{microtype}      

\usepackage{multirow}
\usepackage[multidot]{grffile}
\usepackage{times}
\usepackage{latexsym}
\usepackage{amsmath,amsfonts,amssymb}
\usepackage{graphicx}
\usepackage{subfigure}
\usepackage{colortbl}
\usepackage{algorithm}
\usepackage{algorithmic}

\newcommand{\comment}[1]{}

\newcommand{\mpmi}{\mbox{pmi}}

\newcommand{\DEVELOPMENT}{0} 
\usepackage{ifthen}
\ifthenelse{\DEVELOPMENT = 1}{
			
    \newcommand{\hl}[1]{\textcolor{red}{#1}}
}{
			
    \newcommand{\hl}[1]{#1}		
}

\title{A Simple Language Model based on PMI Matrix Approximations}

  \author{Oren Melamud \\
  IBM Research  \\
  Yorktown Heights, NY, USA  \\
  {\small \tt oren.melamud@ibm.com} \\
  \\\And  
  Ido Dagan \\
  Computer Science Dept.\\
  Bar-Ilan University, Israel\\
 { \small \tt dagan@cs.biu.ac.il} \\
 \\\And
  Jacob Goldberger \\
  Faculty of Engineering \\
Bar-Ilan University, Israel  \\
  {\small \tt jacob.goldberger@biu.ac.il} \\}

\date{}

\begin{document}
\maketitle

\begin{abstract}

In this study, we introduce a new approach for learning language models by training them to estimate word-context pointwise mutual information (PMI), and then deriving the desired conditional probabilities from PMI at test time. 
Specifically, we show that with minor modifications to \emph{word2vec}'s algorithm, we get principled language models that are closely related to the well-established Noise Contrastive Estimation (NCE) based language models.  
A compelling aspect of our approach is that our models are trained
with the same simple negative sampling objective function that is commonly used in \emph{word2vec} to learn word embeddings.

\end{abstract}

\input{introduction}

\input{section2}

\input{experiments}

\section*{Acknowledgments}
This work is supported by the Intel Collaborative Research Institute for Computational Intelligence (ICRI-CI).
\bibliography{paper}
\bibliographystyle{emnlp_natbib}

\end{document}

%% file: introduction.tex
\section{Introduction}

Language models (LMs) learn to estimate the probability of a word
given a context of preceding words.
Recurrent Neural Network (RNN) language models recently outperformed traditional $n$-gram LMs across a range of tasks \cite{jozefowicz2016exploring}.
However, an important
practical issue associated with such neural-network LMs is the high computational
cost incurred. The key factor that limits the scalability of  traditional neural LMs is the computation
of the normalization term in the softmax output layer, whose cost is linearly proportional to the size of the word vocabulary.

Several methods have been proposed to cope with this scaling issue by replacing the softmax with a more computationally efficient component at train time.\footnote{\hl{An alternative recent approach for coping with large word vocabularies is to represent words as compositions of sub-word units, such as individual characters. This approach has notable merits \cite{jozefowicz2016exploring, Sennrich2016NeuralMT}, but is out of the scope of this paper.}} 
These include importance sampling \cite{Bengio2003}, hierarchical softmax \cite{Mnihnips}, BlackOut \cite{ji2016blackout} and Noise Contrastive Estimation (NCE) \cite{Gutmann2012}.
NCE has been applied to train neural LMs with large vocabularies \cite{Mnih2012} and more recently was also successfully used to train LSTM-RNN LMs \cite{Vaswani2013,Chen2015,Zoph2016}. NCE-based language models achieved near state-of-the-art performance on language modeling tasks \cite{jozefowicz2016exploring,Chen2016StrategiesFT}, and as we later show, are closely related to the method presented in this paper.

Continuous word embeddings were initially introduced as a `by-product' of learning neural language models \cite{Bengio2003}. However, they were later adopted in many other NLP tasks, and the most popular recent word embedding learning models are no longer proper language models.
In particular, the skip-gram with negative sampling (NEG) embedding algorithm  \cite{Mikolov_nips} as implemented in the \emph{word2vec} toolkit, has become one of the most popular such models today. This is largely attributed to its scalability to huge volumes of data, which is critical for learning high-quality embeddings.
Recently, \citet{Levy_nips} offered a motivation for the NEG objective function, showing that by maximizing this function, the skip-gram algorithm implicitly attempts to factorize a word-context pointwise mutual information (PMI) matrix. \citet{melamud2017acl} rederived this result by offering an information-theory interpretation of NEG.

The NEG objective function is considered a simplification of the NCE's objective, unsuitable for learning language models \cite{Dyer2014}.
However, in this study, we show that despite its simplicity, it can be used in a principled way to effectively train a language model, based on PMI matrix factorization.
More specifically, we use NEG to train a model for estimating the PMI between words and their preceding contexts, and then derive conditional probabilities from PMI at test time.
The obtained \emph{PMI-LM} can be viewed as a simple variant of \emph{word2vec}'s algorithm, where the context of a predicted word is the preceding sequence of words, rather than a single word within a context window (skip-gram), or a bag-of-context-words (CBOW). 

Our analysis shows that the proposed PMI-LM is very closely related to NCE language models (\emph{NCE-LMs}). 
Similar to NCE-LMs, PMI-LM avoids the dependency of train run-time  on the size of the word vocabulary by sampling from a negative (noise) distribution. Furthermore, conveniently, it also has a notably more simplified objective function formulation inherited from \emph{word2vec}, which allows it to avoid the heuristic components and initialization procedures used in various implementations of NCE language models \cite{Vaswani2013,Chen2015,Zoph2016}.

Finally, we report on a perplexity evaluation of PMI and NCE language models on two standard language modeling datasets. The evaluation yielded comparable results, supporting our theoretical analysis.

%% file: section2.tex
\section{NCE-based Language Modeling}
\label{sec:4}
Noise Contrastive Estimation (NCE) 
has recently been used to learn language models efficiently.
NCE transforms the parameter learning problem into a binary classifier training problem.
Let $p(w|c)$ be the probability of a word $w$ given a context $c$ that represents its entire preceding context, and let $p(w)$ be a `noise'  word distribution (e.g. a unigram distribution). The NCE approach assumes that  the word $w$ is sampled from a mixture distribution $\frac{1}{k\!+\!1}(p(w|c)+kp(w))$ such that the noise samples are $k$ times more frequent than samples from the `true' distribution $p(w|c)$.
Let $y$ be a binary random variable such that $y=0$ and $y=1$ correspond to a noise sample and a true sample, respectively, i.e.
$p(w|c,y=0)=p(w)$ and $p(w|c,y=1)=p(w|c)$.
Assume the distribution $p(w|c)$
has the following parametric form:
\begin{equation}
p_{nce}(w|c) =
\frac{1}{Z_c} \exp(\vec{w} \cdot \vec {c}+b_w)
\label{loglinearm}
\end{equation}
such that $\vec{w}$ and $\vec{c}$ are vector representations of  the word $w$  and its context $c$.
Applying Bayes rule, it can be  easily  verified that:

\begin{equation}
 p_{{nce}}(y=1|w,c)  = \hspace{3cm}
 \label{ncescore}
\end{equation}

$$ \hspace{1cm} \sigma ( \vec{w} \cdot \vec{c} +b_w - \log Z_c  - \log (p(w) k))$$
where $\sigma()$ is the sigmoid function.

NCE uses Eq.~(\ref{ncescore}) and the following objective function to train a binary classifier that decides which distribution was used to sample $w$:
\hl{
\begin{equation} S_{nce}=\sum_{w,c \in D} \Big[  \log p(1|w,c) + \sum_{i=1}^k \log   p(0|u_i,c)\Big]
\end{equation}
}
such that $w,c$ go over all the word-context co-occurrences in the learning corpus $D$ and $u_1,...,u_k$ are `noise' samples drawn from the word unigram distribution.

Note that the normalization factor  $Z_c$ is not a free parameter and to obtain its value,  one needs to compute
 $Z_c=\sum_{w \in V} \exp (\vec{w} \cdot \vec {c}+b_w)$ for each context $c$, where $V$ is the word vocabulary. This computation is typically not feasible due to the large vocabulary size and the exponentially large number of possible contexts and therefore it was heuristically circumvented by prior work.  
\citet{Mnih2012} found empirically
 that setting $Z_c=1$ didn't hurt the performance (see also discussion in \cite{Andreas_2015}).
\citet{Chen2015} reported that setting $\log(Z_c)=9$ gave them the best results.
Recent works \cite{Vaswani2013, Zoph2016} used $Z_{c} = 1$ and also initialized NCE's bias term from Eq. (\ref{ncescore}) to $b_{w} = -\log|V|$.
They reported that without these heuristics the training procedure did not converge to a meaningful model.

In the following section, we describe our proposed language model, which is derived from \emph{word2vec}'s interpretation as a low-rank PMI matrix approximation. 
Interestingly, this model turns out to be a close variant of NCE language models, but with a simplified objective function that avoids the need for the normalization factor $Z_c$ and the bias terms.

\section{PMI-based Language Modeling}
\label{sec:2}
The skip-gram negative sampling word embedding algorithm
represents each word $w$ and each context word $c$ as $d$-dimensional vectors, with the purpose that words that are ``similar" to each other  will have similar vector representations.
The algorithm optimizes the following NEG objective function \cite{Mikolov_nips}:
\hl{
\begin{equation} S_{neg} = \sum_{w,c \in D} \Big[ \log \sigma (\vec{w} \cdot \vec{c}) + \sum_{i=1}^k  \log \sigma (-\vec{u}_i \cdot \vec{c})\Big]
\label{eq:negobj}
\end{equation}
}
\noindent such that $w,c$ go over all the word-context co-occurrences in the learning corpus $D$, $u_1,...,u_k$ are words independently sampled from the word unigram distribution, $\vec{x}$ is the embedding of $x$ and $\sigma()$ is the sigmoid function.
\hl{The objective function $S_{neg}$ can be
viewed as a log-likelihood function of a binary logistic
regression classifier that treats a sample from
a joint word-context distribution as a positive instance,
and two independent samples from the word and context unigram
distributions as a negative instance}, while $k$ is the
proportion between negative and positive instances.
\citet{Levy_nips} showed that this objective function achieves its maximal value when for every word-context pair~$w,c$:
\begin{equation}  \vec{w} \cdot \vec{c} = \mpmi_{k}(w,c) = \log \frac{p(w|c)}{kp(w)}
\label{pmi}
\end{equation}
\hl{ \noindent where $\mpmi_{k}(w,c)$ is the word-context \emph{PMI matrix}.
Actually achieving this maximal value is typically infeasible, since the embedding dimensionality is intentionally limited. Therefore, learning word and context embeddings that optimize skip-gram's NEG objective function
(\ref{eq:negobj}) can be viewed as finding a low-rank approximation of the word-context PMI matrix.}
An explicit expression of the approximation criterion optimized by the skip-gram algorithm can be found in  \cite{melamud2017acl}.

Our study is based on two simple observations regarding this finding of \citet{Levy_nips}.
First, Equation (\ref{pmi}) can be reformulated as follows to derive an estimate of the conditional distribution $p(w|c)$:
 \begin{equation}
\hat{p}(w|c) \propto  \exp(  \vec{w} \cdot \vec{c} ) p(w)
\label{conditional}
\end{equation}
where the constant $k$ is dropped since $p(w|c)$ is a distribution.
Second,
while the above analysis had been originally applied to the case of word-context joint distributions $p(w,c)$, it is easy to see that the PMI matrix approximation analysis also holds for every Euclidean embedding of a joint distribution $p(x,y)$ of any two given random variables $X$ and $Y$.
In particular, we note that it holds for word-context joint distributions $p(w,c)$, where $w$ is a single word, but $c$ represents its entire preceding context, rather than just a single context word, and $\vec{c}$ is a vector representation of this entire context.
\hl{Altogether, this allows us to use \emph{word2vec}'s NEG objective function (\ref{eq:negobj}) to approximate the language modeling conditional probability $\hat{p}(w|c)$ (\ref{conditional}),
with~$c$ being the entire preceding context of the predicted word $w$.}

 \begin{table*}[t]
\center

\begin{tabular}{|l|l|l|}
  \hline
   & Training objective function & Test probability estimate\\\hline
   NCE-LM & $s(w,c)=\sigma(\vec{w} \cdot \vec{c} + b_w \!-\! \log Z_c\!- \! \log (kp(w))) $ & $\hat p(w|c) \propto \exp (\vec{w} \cdot \vec{c} +b_w) $ \\
   PMI-LM & $s(w,c)=\sigma(\vec{w} \cdot\vec{c}) $ &  $\hat p(w|c) \propto \exp (\vec{w} \cdot \vec{c}) p(w) $  \\
  \hline
\end{tabular}
\caption{Comparison of the training objective functions \hl{(see Eq. (\ref{eq:common_obj}))}
and the respective test-time conditional
word probability functions for NCE-LM and PMI-LM algorithms. 
}
\label{tab:strategies}
\end{table*}

We next describe the \hl{design} details of the proposed PMI-based language modeling.
We use a simple lookup table for the word representation $\vec{w}$, and an  LSTM recurrent neural network
to obtain a low dimensional representation of the entire preceding context~$\vec{c}$.
These representations are trained to maximize the NEG objective in Eq.~(\ref{eq:negobj}), where this time $w$ goes over every word token in the corpus, and $c$ is its preceding context.
We showed above that optimizing this objective seeks to obtain the best low-dimensional approximation of the PMI matrix associated with the joint distribution of the word and its preceding context (Eq.~(\ref{pmi})).
Hence, based on Eq.~(\ref{conditional}), for a reasonable embedding dimensionality and a good model for representing the preceding context, we expect $\hat{p}(w|c)$ to be a good estimate of the language modeling conditional distribution.

At test time, to obtain a proper distribution, we perform a normalization operation as done by all other comparable models.
The train and test steps of the proposed language modeling algorithm  are shown in algorithm box~\ref{labelR}.

\begin{algorithm}[tb]
   \caption{ PMI Language Modeling }
     \label{labelR}
\begin{algorithmic}
\STATE \STATE \textbf{Training phase}:

    \STATE - Use a simple lookup table for the word representation and an LSTM recurrent neural network
to obtain the preceding context representation.
\STATE - Train the word and preceding context embeddings to maximize the objective:
$$ S_{neg} = \sum_{w,c \in D} \Big[ \log \sigma (\vec{w} \cdot \vec{c}) + \sum_{i=1}^k  \log \sigma (-\vec{u}_i \cdot \vec{c})\Big]
$$
such that $w$ and $c$ go over every word and it preceding context in the corpus $D$, and $u_1,...,u_k$ are words independently  sampled from   the unigram distribution $p(w)$.
\STATE
\vspace{-5 pt}
\STATE \textbf{Test phase}:
\STATE The conditional probability estimate for a word~$w$ given a preceding context~$c$ is:
   $$
\hat{p}(w|c) = \frac{   \exp(  \vec{w} \cdot \vec{c} )p(w)} { \sum_{v \in V}  \exp(  \vec{v} \cdot \vec{c} )p(v)}
$$
where $V$ is the word vocabulary.
   \end{algorithmic}
\end{algorithm}

Note that while the NCE approach (\ref{loglinearm}) learns to explicitly estimate normalized conditional distributions, 
our model learns to approximate the PMI matrix. Hence, we have no real motivation to include additional learned normalization parameters, as considered in comparable NCE language models \cite{Mnih2012,Zoph2016}.

The \hl{NEG} and NCE objective functions share a similar form:
\hl{
\begin{equation}
S = \sum_{w,c} \Big[ \log s(w,c) + \sum_{i=1}^k  \log (1-s(u_i,c)) \Big]
\label{eq:common_obj}
\end{equation}
}
with the differences summarized in Table~\ref{tab:strategies}.
The comparison shows that PMI-LM's \hl{NEG} objective function is much simpler.  Furthermore, due to the component $\log (p(w) k))$ in NCE's objective function, its input to the sigmoid function is sensitive to the variable values in the unigram distribution, and therefore potentially more difficult to concentrate around zero with low variance to facilitate effective back-propagation. This may explain heuristics used by prior work for initializing the values of $b_w$ \cite{Vaswani2013,Zoph2016}.

%% file: experiments.tex
\section{Experiments}
\label{sec:5}

\label{subsec:lm}
The goal of the evaluation described in this section is to empirically establish PMI-LM as a sound language model. We do so by comparing its performance with the well-established NCE-LM, using the popular perplexity measure on two standard datasets, under the same terms.  
We describe our hyperparameter choices below and stress that for a fair comparison, we followed prior best practices and avoided hyperparameter optimization in favor of PMI-LM.
All of the models described hereafter were implemented using the Chainer toolkit \cite{tokuichainer}.

For our NCE baseline, we used the heuristics that worked well in \cite{Vaswani2013,Zoph2016}, initializing NCE's bias term from Eq. (\ref{ncescore}) to $b_{w} = -\log|V|$, where $V$ is the word vocabulary, and using $Z_{c} = 1$. 

The first dataset we used is a version of the Penn Tree Bank (PTB), commonly used to evaluate language models.\footnote{Available from Tomas Mikolov at: \url{http://www.fit.vutbr.cz/~imikolov/rnnlm/simple-examples.tgz}} It consists of 929K training words, 73K validation words and 82K test words with a 10K
word vocabulary. To build and train the compared models in this setting, we followed the work of Zaremba et al. \shortcite{zaremba2014recurrent}, who achieved excellent results on this dataset. Specifically, we used a 2-layer 300-hidden-units LSTM with a 50\% dropout ratio to represent the preceding (left-side) context of a predicted word.\footnote{Zaremba et al. \shortcite{zaremba2014recurrent} used larger models with more units and also applied dropout to the output of the top LSTM layer, which we did not.} We represented end-of-sentence as a special $<$eos$>$ token and predicted this token like any other word. During training, we performed truncated back-propagation-through-time, unrolling the LSTM for 20 steps at a time without ever resetting the LSTM state. We trained our model for 39 epochs using Stochastic Gradient Descent (SGD) with a learning rate of 1, which is decreased by a factor of 1.2 after every epoch starting after epoch~6. We clipped the norms of the gradient to 5 and used a mini-batch size of 20. We set the negative sampling parameter to $k=100$ following Zoph et al. \shortcite{Zoph2016}, who showed highly competitive performance with NCE LMs trained with this number of samples.

As the second dataset, we used the much larger WMT 1B-word benchmark introduced by Chelba et al. \shortcite{chelba2013one}. This dataset comprises about 0.8B training words and has a held-out set partitioned into 50 subsets. The test set is the first subset in the held-out, comprising 159K words, including the $<$eos$>$ tokens. We used the second subset as the validation set with 165K words. The original vocabulary size of this dataset is 0.8M words after converting all words that occur less than 3 times in the corpus to an $<$unk$>$ token. However, we followed previous works \cite{williams2015scaling,ji2016blackout} and trimmed the vocabulary further down to the top 64K most frequent words in order to successfully fit a neural model to this data using reasonably modest compute resources. To build and train our models, we used a similar method to the one used with PTB, with the following differences. We used a single-layer 512-hidden-unit LSTM to represent the preceding context. We followed Jozefowicz et al. \shortcite{jozefowicz2016exploring}, who found a 10\% dropout rate to be sufficient for relatively small models fitted to this large training corpus. We trained our model for only one epoch using the Adam optimizer \cite{kingma2014adam} with default parameters, which we found to converge more quickly and effectively than SGD. We used a mini-batch size of 1000.

\begin {table}[t]
\begin{center}

\begin{tabular}{|l|c|r|r|}
\hline
	 & PMI-LM &   NCE-LM \\
\hline
	PTB & \textbf{98.35}  & 104.33 \\
\hline
	WMT & \textbf{65.84}  &  69.28 \\
\hline
\end{tabular}
\end{center}

\caption{Perplexity results on test sets.}
\label{tab:results}
\end {table}

The perplexity results 
achieved by the compared models 
appear in Table \ref{tab:results}. 
As can be seen, the performance of our PMI-LM is competitive, slightly outperforming the NCE-LM on both test sets.  
To put these numbers in a broader context, we note that state-of-the-art results on these datasets are notably better. For example, 
on the small PTB test set, \citet{zaremba2014recurrent} achieved 78.4 perplexity with a larger LSTM model and using the more costly softmax component. \hl{On the larger WMT dataset, \citet{jozefowicz2016exploring} achieved
46.1 and 43.7 perplexity numbers using NCE and importance sampling respectively, and with much larger LSTM models trained over the full vocabulary, rather than our trimmed one. They also achieved 23.7 with an ensemble method, which is the best result on this dataset to date.} Yet, as intended, we argue that our experimental results affirm the claim that PMI-LM is a sound language model on par with NCE-LM.

\section{Conclusions}

In this work, we have shown that \emph{word2vec}'s negative sampling objective function, popularized in the context of learning word representations, can also be used to effectively learn parametric language models. These language models are closely related to NCE language models, but utilize a simpler, potentially more robust objective function.
More generally, our theoretical analysis shows that any \emph{word2vec} model trained with negative sampling can be used in a principled way to estimate the conditional distribution $p(w|c)$, by following our proposed procedure at test time.